\documentclass[letterpaper, 10 pt, conference]{ieeeconf}  
\pdfoutput=1

\IEEEoverridecommandlockouts                              

\usepackage{graphics} 
\usepackage{epsfig} 
\usepackage{mathptmx} 
\usepackage{times} 
\usepackage{amsmath} 
\usepackage{amssymb}  
\usepackage{multirow}
\usepackage{array}
\usepackage{booktabs}
\usepackage{caption}
\usepackage{subfigure}

\title{\LARGE \bf
StarNet: Pedestrian Trajectory Prediction using \\
Deep Neural Network in Star Topology
}

\author{Yanliang Zhu, Deheng Qian, Dongchun Ren, Huaxia Xia
\thanks{*This work was supported by the Meituan-Dianping Group.}
\thanks{Yanliang Zhu, Deheng Qian, Dongchun Ren and Huaxia Xia are with the Meituan-Dianping Group, Beijing, China.
        {\tt\small zhuyanliang@meituan.com}}%
}

\begin{document}

\maketitle
\thispagestyle{empty} 



\begin{abstract}

Pedestrian trajectory prediction is crucial for many important applications. This problem is a great challenge because of complicated interactions among pedestrians. Previous methods model only the pairwise interactions between pedestrians, which not only oversimplifies the interactions among pedestrians but also is computationally inefficient. In this paper, we propose a novel model StarNet to deal with these issues. StarNet has a star topology which includes a unique hub network and multiple host networks. The hub network takes observed trajectories of all pedestrians to produce a comprehensive description of the interpersonal interactions. Then the host networks, each of which corresponds to one pedestrian, consult the description and predict future trajectories. The star topology gives StarNet two advantages over conventional models. First, StarNet is able to consider the collective influence among all pedestrians in the hub network, making more accurate predictions. Second, StarNet is computationally efficient since the number of host network is linear to the number of pedestrians. Experiments on multiple public datasets demonstrate that StarNet outperforms multiple state-of-the-arts by a large margin in terms of both accuracy and efficiency.

\end{abstract}

\section{INTRODUCTION}

Pedestrian trajectory prediction is an important task in autonomous driving \cite{c1,c2,c3} and mobile robot applications \cite{c4,c5,c6}. This task allows an intelligent agent, e.g., a self-driving car or a mobile robot, to foresee the future positions of pedestrians. Depending on such predictions, the agent can move in a safe and smooth route.

However, pedestrian trajectory prediction is a great challenge due to the intrinsic uncertainty of pedestrians' future positions. In a crowded scene, each pedestrian dynamically changes his/her walking speed and direction, partly attributed to his/her interactions with surrounding pedestrians.

\begin{figure}[thpb]
\centering
\includegraphics[width=8cm]{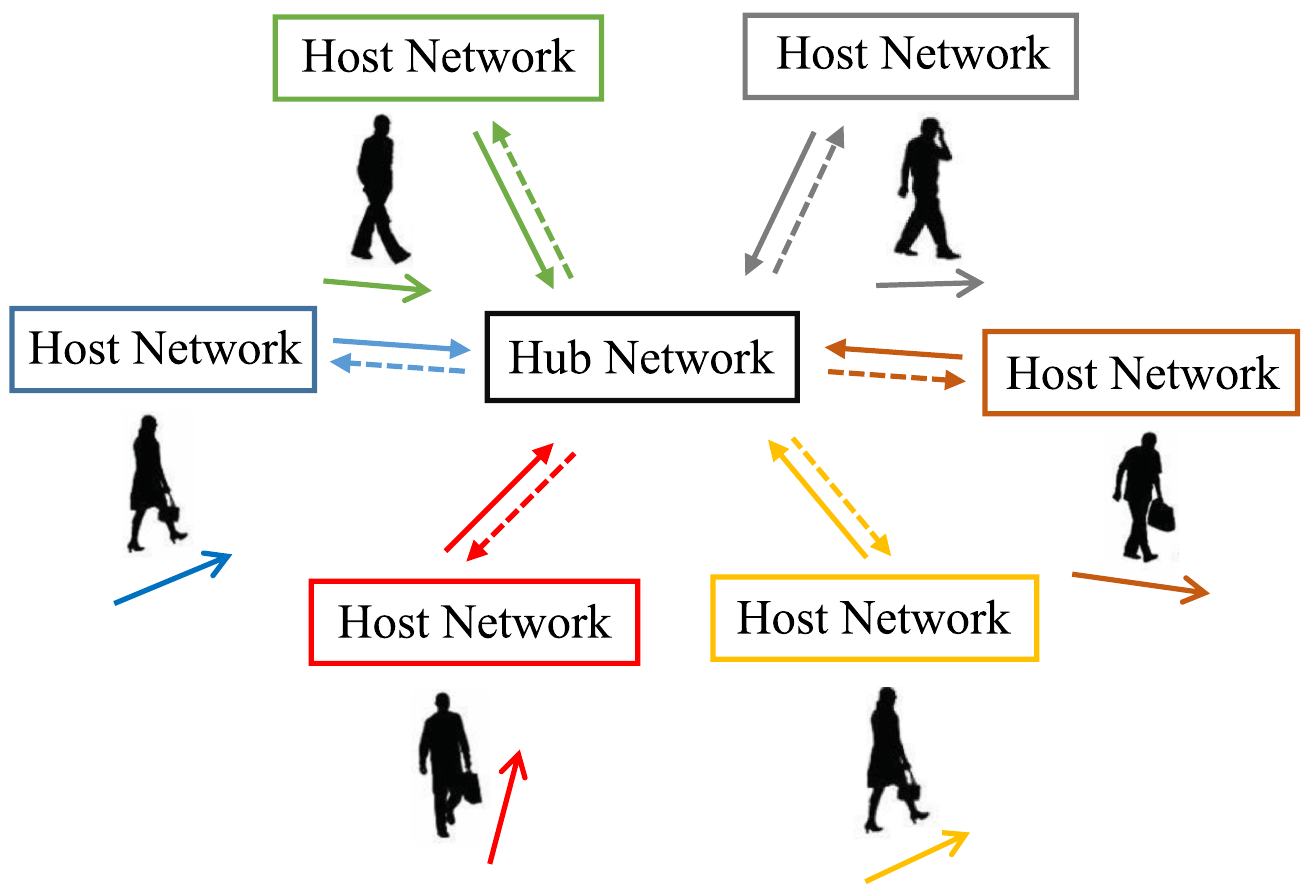}
\caption{The structure of StarNet. StarNet mainly consists a centralized hub network and several host networks. The hub network collects movement information and generates a feature which describes joint interactions among pedestrians. Each host network, corresponding to a certain pedestrian, queries the hub network and predicts the pedestrian's trajectory.}
\label{fig:1}
\end{figure}

To make an accurate prediction, existing algorithms focus on making full use of the interactions between pedestrians. Early works model the interactions \cite{c7,c8,c9,c10} by hand-crafted features. Social Force \cite{c7} models several force terms to predict human behaviors. The approach in \cite{c8} constructs an energy grid map to describe the interactions in crowded scenes. However, their performances are limited by the quality of manually designed features. Recently, data-driven methods have demonstrated their powerful performance \cite{c11,c12,c13,c14}. For instance, Social LSTM \cite{c11} considers interactions among pedestrians close to each other. Social GAN \cite{c13} models interactions among all pedestrians. Social Attention \cite{c14} captures spatio-temporal interactions.

Previous methods have achieved great success in trajectory prediction. However, all these methods assume that the complicated interactions among pedestrians can be decomposed into pairwise interactions. This assumption neglects the collective influence among pedestrians in the real world. Thus previous methods tend to fail in complicated scenes. In the meanwhile, the number of pairwise interactions increases quadratically as the number of pedestrians increases. Hence, existing methods are computationally inefficient.

In this paper, we propose a new deep neural network, StarNet, to model complicated interactions among all pedestrians together. As shown in Figure \ref{fig:1}, StarNet has a star topology, and hence the name. The central part of StarNet is the hub network, which produces a representation $\mathbf{r}$ of the interactions among pedestrians. To be specific, the hub network takes the observed trajectories of all pedestrians and produces a comprehensive spatio-temporal representation $\mathbf{r}$ of all interactions in the crowd. Then, $\mathbf{r}$ is sent to each host network. Each host network predicts one pedestrian's trajectory. Specifically, depending on $\mathbf{r}$, each host network exploits an efficient method to calculate the pedestrian's interactions with others. Then, the host network predicts one pedestrian's trajectory depending on his/her interactions with others, as well as his/her observed trajectory.

StarNet has two advantages over previous methods. First, the representation $\mathbf{r}$ is able to describe not only pairwise interactions but also collective ones. Such a comprehensive representation enables StarNet to make accurate predictions. Second, the interactions between one pedestrian and others are efficiently computed. When predicting all pedestrians' trajectories, the computational time increases linearly, rather than quadratically, as the number of pedestrians increases. Consequently, StarNet outperforms multiple state-of-the-arts in terms of both accuracy and computational efficiency.

Our contributions are two-folded. First, we propose to describe collective interactions among pedestrians, which results in more accurate predictions. Second, we devise an interesting topology of the network to take advantage of the representation $\mathbf{r}$, leading to computational efficiency.

The rest of this paper is organized as follows: Section \ref{section:2} briefly reviews related work on pedestrian trajectory prediction. Section \ref{section:3} formalizes the problem and elaborates our method. Section \ref{section:4} compares StarNet with state-of-the-arts on multiple public datasets. Section \ref{section:5} draws our conclusion.

\section{RELATED WORK}\label{section:2}

Our work mainly focuses on human path prediction. In this section, we give a brief review of recent researches on this domain.

Pedestrian path prediction is a great challenge due to the uncertainty of future movements \cite{c7,c8,c10,c11,c13,c14,c15}. Conventional methods tackle this problem with manually crafted features. Social Force \cite{c7} extracts force terms, including self-properties and attractive effects, to model human behaviors. Another approach \cite{c8} constructs an energy map to indicate the traffic capacity of each area in the scene, and uses a fast matching algorithm to generate a walking path. Mixture model of Dynamic pedestrian-Agents (MDA) \cite{c10} learns the behavioral patterns by modeling dynamic interactions and pedestrian beliefs. However, all these methods can hardly capture complicated interactions in crowded scenes, due to the limitation of hand-crafted features.

Data-driven methods remove the requirement of hand-crafted features, and greatly improve the ability to predict pedestrian trajectories. Some attempts \cite{c11,c13,c14,c26,c27} receive pedestrian positions and predict determined trajectories. Social LSTM \cite{c11} devises social pooling to deal with interpersonal interactions. Social LSTM divides pedestrian's surrounding area into grids, and computes pairwise interactions between pedestrians in a grid. Compared with Social LSTM, other approaches \cite{c13,c15} eliminate the limitation on a fixed area. Social GAN \cite{c13} combines Generative Adversarial Networks (GANs) \cite{c16} with LSTM-based encoder-decoder architecture, and sample plausible trajectories from a distribution. Social Attention \cite{c14} estimates multiple Gaussian distributions of future positions, then generates candidate trajectories through Mixture Density Network (MDN) \cite{c17}.

However, existing methods compute pairwise features, and thus oversimplified the interactions in the real word environment. Meanwhile, they suffer from a huge computational burden in crowded scenes. In contrast, our proposed StarNet with novel architecture is capable of capturing joint interactions over all pedestrians, which is more accurate and efficient.

\section{APPROACH}\label{section:3}

In this section, we first describe the formulation of the pedestrian prediction problem. Then we provide the details of our proposed method.

\begin{figure*}[thpb]
\centering
\includegraphics[width=13cm]{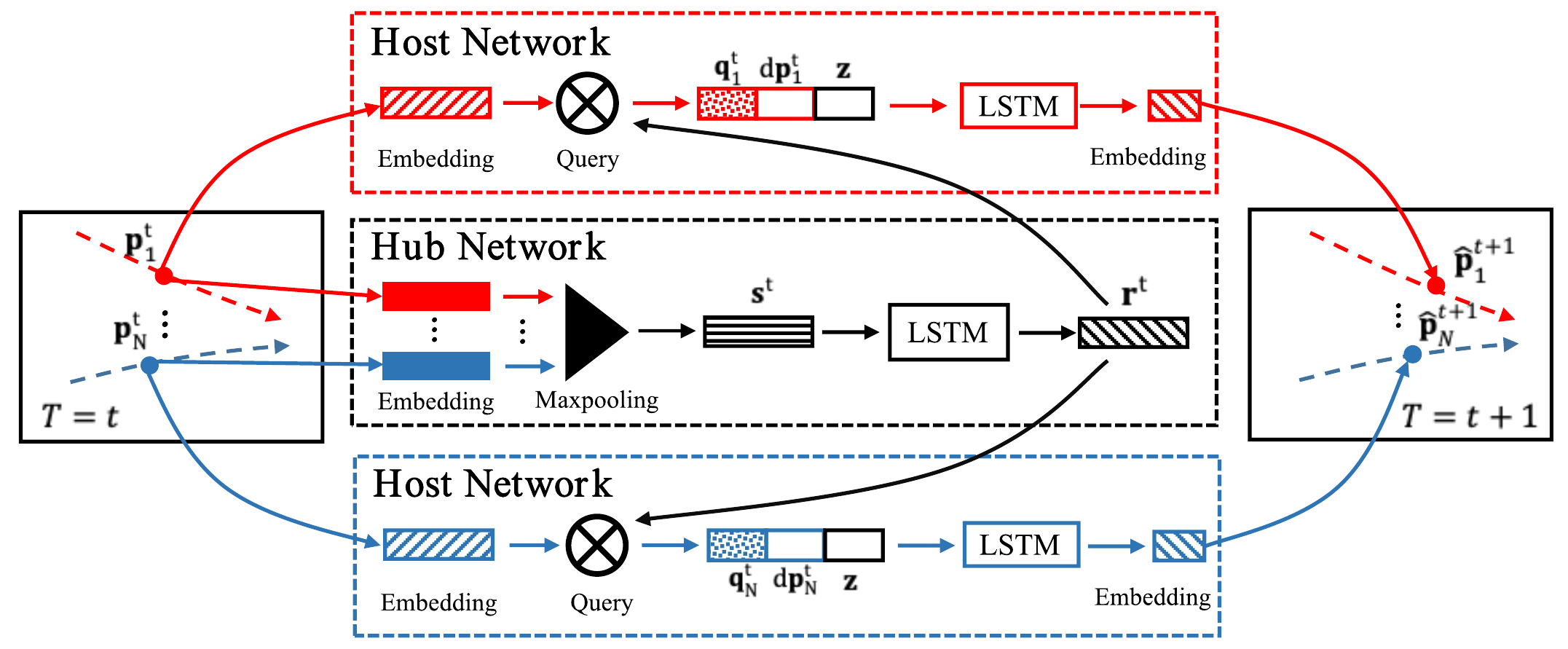}
\caption{The process of predicting the coordinates. At time step $t$, StarNet takes the newly observed (or predicted) coordinates $\left\{\mathbf{p}_i^t\right\}_{i=1}^N$ (or $\left\{\mathbf{\widehat{p}}_i^t\right\}_{i=1}^N$) and outputs the predicted coordinates $\left\{\mathbf{\widehat{p}}_i^{t+1}\right\}_{i=1}^N$.}
\label{fig:2}
\end{figure*}

\subsection{Problem Formulation}

We assume the number of pedestrians is $N$. The number of observed time steps is $T_{obs}$. And the number of time steps to be predicted is $T_{pred}$. For the $i$-th pedestrian, his/her observed trajectory is denoted as $O_{i} = \left\{\mathbf{p}_{i}^{t} \mid t=1,2,\cdots,T_{obs}\right\}$, where $\mathbf{p}_{i}^{t}$ represents his/her coordinates at time step $t$. Similarly, the future trajectory of ground truth is denoted as $F_{i} = \left\{\mathbf{p}_{i}^{t} \mid t=T_{obs}+1,T_{obs}+2,\cdots,T_{obs}+T_{pred}\right\}$.

Given such notations, our goal is to build a fast and accurate model to predict the future trajectories $\left\{F_{i}\right\}_{i=1}^N$ of all pedestrians, based on their observed trajectories $\left\{O_{i}\right\}_{i=1}^N$. In other words, we try to find a function mapping from $\left\{O_{i}\right\}_{i=1}^N$ to $\left\{F_{i}\right\}_{i=1}^N$. We employ a deep neural network, which is called StarNet, to embody this function. Specifically, StarNet consists of two novel parts, i.e., a hub network and $N$ host networks. The hub network computes a representation $\mathbf{r}$ of the crowd. Then, each host network predicts the future trajectory of one pedestrian depending on the pedestrian's observed trajectory and $\mathbf{r}$. We first describe the hub network and then present host networks.

\subsection{The hub network}

The hub network takes all of the observed trajectories simultaneously and produces a comprehensive representation $\mathbf{r}$ of the crowd of pedestrians. The representation $\mathbf{r}$ includes both spatial and temporal information of the crowd, which is the key to describe the interactions among pedestrians.

Note that our algorithm should be invariant against isometric transformation (translation and rotation) of the pedestrians' coordinates. The invariance against rotation is achieved by randomly rotate our training data during the training process. While the invariance against translation is guaranteed by calculating a translation invariant representation $\mathbf{r}$.

As shown in Figure \ref{fig:2}, the hub network produces $\mathbf{r}$ by two steps. First, the hub network produces a spatial representation of the crowd for each time step. The spatial representation is invariant against the translation of the coordinates. Then, the spatial representation is fed into a LSTM to produce the spatio-temporal representation $\mathbf{r}$.

\subsubsection{Spatial representation}
In the first step, in order to make the representation invariant against translation, the hub network preprocesses the coordinates of pedestrians by subtracting the central coordinates of all pedestrians at time step $T_{obs}$ from every coordinate.
$$
\mathbf{p}_{i}^{t} \leftarrow \mathbf{p}_{i}^{t} - \frac{1}{N}\sum_{n=1}^{N} \mathbf{p}_{n}^{T_{obs}}. \eqno{(1)}
$$

Thus, the centralized coordinates are invariant against translation. Such coordinates of each pedestrian are mapped into a new space using an embedding function $\phi(\cdot)$ with parameters $W_{1}$,
$$
\mathbf{e}_{i}^{t} = \left\{
\begin{array}{ll}
\phi \left(\mathbf{p}_{i}^{t};W_{1} \right), & if\ t \in \left[1,T_{obs}\right], \\
\phi \left(\widehat{\mathbf{p}}_{i}^{t};W_{1} \right), & if\ t \in \left[T_{obs}+1, T_{obs}+T_{pred}\right],
\end{array}
\right. \eqno{(2)}
$$
where $\widehat{\mathbf{p}}_i^t$ is the predicted position of the $i$-th pedestrian at time step $t$. $\mathbf{e}_{i}^{t}$ is the spatial representation of the $i$-th pedestrian's trajectory at time step $t$. The embedding function is defined as:
$$
\phi \left(\mathbf{x};W \right) \triangleq W \mathbf{x}.  \eqno{(3)}
$$

Then, we use a maxpooling operation to combine the spatial representation of all pedestrians, obtaining the spatial representation of the crowd at time step $t$,
$$
\mathbf{s}^{t} = MaxPooling \left(\mathbf{e}_{1}^{t},\mathbf{e}_{2}^{t},\cdots,\mathbf{e}_{N}^{t} \right), \eqno{(4)}
$$

Spatial representation $\mathbf{s}^t$ contains information of the crowd at a single time step. However, pedestrians interact with each other dynamically. To improve the accuracy of predictions, a spatio-temporal representation is required.

\subsubsection{Spatio-temporal representation}
In the second step, the hub network feeds a set of spatial representations $\left\{\mathbf{s}^1,\mathbf{s}^2,\cdots,\mathbf{s}^{T_{obs}}\right\}$ of sequential time steps into a LSTM. Then, the LSTM combines all the spatial representations in its hidden state. Thus, the hidden state of the LSTM is a spatio-temporal representation $\mathbf{r}^t$ of all pedestrians. Specifically, we can calculate $\mathbf{r}^t$ as follows:
$$
\left\{
\begin{array}{l}
\mathbf{h}_{c}^{0}=\mathbf{0}, \\
\mathbf{e}^{t} = \phi \left(\mathbf{s}^{t};W_{2} \right), \\
\left[\mathbf{o}_{c}^{t}, \mathbf{h}_{c}^{t}\right]= LSTM \left(\mathbf{h}_{c}^{t-1}, \mathbf{e}^{t}; W_{3} \right),\\
\mathbf{r}^{t} = \phi \left(\mathbf{o}_{c}^{t};W_{4} \right),
\end{array}
\right. \eqno{(5)}
$$
where $W_{2}$ and $W_{4}$ are the embedding weights, $W_{3}$ is the weight of LSTM. $\mathbf{o}_{c}^{t}$ and $\mathbf{h}_{c}^{t}$ are the output and hidden state of the LSTM respectively.

Note that, $\mathbf{r}^t$ depends on the observed trajectories of all pedestrians. Hence, our algorithm is able to consider complicated interactions among multiple pedestrians. This property allows our algorithm to produce accurate predictions. Meanwhile, $\mathbf{r}^t$ is able to be obtained in a single forward propagation of the hub network at each time step. In other words, the time complexity of computing interactions among pedestrians is linear to the number of pedestrians $N$. This property allows our algorithm to be computationally efficient. By contrast, conventional algorithms compute pairwise interactions, leading to oversimplification of the interactions among pedestrians. Also, the number of pairwise interactions increases quadratically as $N$ increases.

\subsection{The host networks}

The spatio-temporal representation $\mathbf{r}^t$ is then employed by host networks. For the $i$-th pedestrian, the host network first embeds the observed trajectory $O_i$, and then combines the embedded trajectory with the spatio-temporal representation $\mathbf{r}^t$, predicting the future trajectory. Specifically, the host network predicts the future trajectory by two steps.

First, the host network takes the observed trajectory $O_i$ and the spatio-temporal representation $\mathbf{r}^t$ as input and generates an integrated representation $\mathbf{q}_i^t$,
$$
\mathbf{q}_{i}^{t} = \left\{
\begin{array}{ll}
\mathbf{r}^{t} \odot \phi\left(\mathbf{p}_{i}^{t};W_{5}\right),&if\ t \in \left[1,T_{obs}\right], \\
\mathbf{r}^{t} \odot \phi\left(\widehat{\mathbf{p}}_{i}^{t};W_{5}\right),&if\ t \in \left[T_{obs}+1,T_{obs}+T_{pred}\right], 
\end{array}
\right. \eqno{(6)}
$$
where $W_{5}$ is the embedding weight, and $\odot$ denotes the point-wise multiplication. $\mathbf{q}_{i}^{t}$ depends on both the trajectory of the $i$-th pedestrian and the interactions between the $i$-th pedestrian and others in the crowd.

Second, the host network predicts the future trajectory of the $i$-th pedestrian depending on the observed trajectory $O_i$ and the integrated representation $\mathbf{q}_i^t$. To encourage the host network to produce non-deterministic predictions, a random noise $\mathbf{z}$, which is sampled from a Gaussian distribution with mean 0 and variance 1, is concatenated to the input of the host network. Specifically, the host network encodes the observed trajectory $O_i$ with the hidden state $\mathbf{h}_{ei}^{T_{obs}}$, i.e.,
$$
\left\{
\begin{array}{ll}
\mathrm{d}\mathbf{p}_{i}^{0} = \mathbf{0}, & \\
\mathrm{d}\mathbf{p}_{i}^{t-1} = \mathbf{p}_{i}^{t} - \mathbf{p}_{i}^{t-1}, &\\
\left[\mathbf{o}_{ei}^{t}, \mathbf{h}_{ei}^{t}\right] = LSTM_{E} \left(\mathbf{h}_{ei}^{t-1},\left[\mathbf{q}_{i}^{t},\mathrm{d}\mathbf{p}_{i}^{t-1} \right];W_{6} \right), & \\
t \in [1, T_{obs}],
\end{array}
\right. \eqno{(7)}
$$
where $LSTM_{E}(\cdot)$ with weight $W_{6}$ denotes the encoding procedure. Then, the host network proceeds with 
$$
\left\{
\begin{array}{ll}
\left[\mathbf{o}_{di}^{t}, \mathbf{h}_{di}^{t}\right] = LSTM_{D} \left(\mathbf{h}_{di}^{t-1}, \left[\mathbf{q}_{i}^{t}, \mathrm{d}\mathbf{\widehat{p}}_{i}^{t-1}, \mathbf{z} \right];W_{7} \right),\\
\mathrm{d}\mathbf{\widehat{p}}_{i}^{t} = \phi\left(\mathbf{o}_{di}^{t};W_{8}\right), \\
\mathbf{\widehat{p}}_{i}^{t} = \mathbf{\widehat{p}}_{i}^{t-1} + \mathrm{d}\mathbf{\widehat{p}}_{i}^{t}, \\
t \in [T_{obs}+1, T_{obs}+T_{pred}],
\end{array}
\right. \eqno{(8)}
$$
where $LSTM_{D}(\cdot)$ with weight $W_{7}$ is the decoding function. $W_{8}$ is the embedding weight of the output layer. And the initial states are set according to,
$$
\left\{
\begin{array}{ll}
\mathbf{h}_{di}^{T_{obs}} = \mathbf{h}_{ei}^{T_{obs}}, & \\
\widehat{\mathbf{p}}_{i}^{T_{obs}} =  \mathbf{p}_{i}^{T_{obs}}, & \\
\mathrm{d}\widehat{\mathbf{p}}_{i}^{T_{obs}} =  \mathbf{p}_{i}^{T_{obs}}-\mathbf{p}_{i}^{T_{obs}-1}.\\
\end{array}
\right. \eqno{(9)}
$$

\subsection{Implementation Details}
The network configuration of StarNet is detailed in TABLE \ref{table:1}.
\begin{table}[h]
\caption{Network Configuration of AstoridNet}
\label{table:1}
\begin{center}
\renewcommand\arraystretch{1.3}
\begin{tabular}{p{2.5cm}<{\centering}|p{4.5cm}<{\centering}}
\hline
$\bf{Weight}$ & $\bf{Weight\ Dimension}$ \\
\hline
$W_{1}$ & 64x2 \\
\hline
$W_{2}$ & 64x64 \\
\hline
$W_{3}$ & 64x32,\ 32x1(bias) \\
\hline
$W_{4}$ & 32x64 \\
\hline
$W_{5}$ & 64x2 \\
\hline
$W_{6}$ & 64x66,\ 64x1(bias) \\
\hline
$W_{7}$ & 64x74,\ 64x1(bias) \\
\hline
$W_{8}$ & 2x64 \\
\hline

\end{tabular}
\end{center}
\end{table}

We train the proposed StarNet with the loss function applied in \cite{c13}. Specifically, at the training stage, StarNet produces multiple predicted trajectories for each pedestrian. Each predicted trajectory $\{\widehat{F}_{ik}\}_{k=1}^{K}$ has a distance to the ground truth trajectory $F_i$. Only the smallest distance is minimized. Mathematically, the loss function is,
$$
L = \frac{1}{NT_{pred}} min_{k=1}^{K}\sum_{j=1}^{N}\sum_{t=T_{obs}+1}^{T_{obs}+T_{pred}} \left(\mathbf{\widehat{p}}_{jk}^{t}-\mathbf{p}_{j}^{t} \right)^{2}, \eqno{(10)} 
$$
where $K$ is the number of sampled trajectories. This loss function improves the training speed and stability. Moreover, we employ an Adam optimizer and set the learning rate to $0.0001$. 

In practice, all host networks share the same weights, since pedestrians in a scenario have the same behavioral patterns, such as variable-speed movement, sharp turning and so on. In our approach, we use shared weights to learn the aforementioned behavioral patterns. Each host network contains specific LSTM state which captures certain pedestrian’s behavior,  and predicts the pedestrian's future trajectory. The observed trajectories of all pedestrians form a batch, which is fed into one single implementation of the host network. In this way, the prediction for all pedestrians is able to be obtained in a single forward propagation.

\begin{table*}[h]
\caption{Comparison of Prediction Errors}
\label{table:2}
\centering
\renewcommand\arraystretch{1.3}
\begin{tabular}{p{2.5cm}<{\centering}|p{1.5cm}<{\centering}|p{1.5cm}<{\centering}|p{1.7cm}<{\centering}|p{2.0cm}<{\centering}|p{2.5cm}<{\centering}|p{2.5cm}<{\centering}}
\hline
$\bf{Metric}$ & $\bf{Dataset} $ & $\bf{LSTM}$ & $\bf{Social\ LSTM}$ & $\bf{Social\ GAN}$ & $\bf{Social\ Attention}$ & $\bf{StarNet\ (Ours)}$ \\
\hline
\multirow{5}{*}{$\bf{ADE}$} & $\bf{ZARA}$-$\bf{1}$ & 0.25 & 0.27 & $\bf{0.21}$ & 1.66 & 0.25 \\
\cline{2-7}
~ & $\bf{ZARA}$-$\bf{2}$ & 0.31 & 0.33 & 0.27 & 2.30 & $\bf{0.26}$ \\
\cline{2-7}
~ & $\bf{UNIV}$ & 0.36 & 0.41 & 0.36 & 2.92 & $\bf{0.21}$\\
\cline{2-7}
~ & $\bf{ETH}$ & 0.70 & 0.73 & 0.61 & 2.45 & $\bf{0.31}$\\
\cline{2-7}
~ & $\bf{HOTEL}$ & 0.55 & 0.49 & 0.48 & 2.19 & $\bf{0.46}$\\
\hline
$\bf{Average\ ADE}$ & -  & 0.43 & 0.45 & 0.39 & 2.30 & $\bf{0.30}$\\
\hline
$\bf{Variance\ of\ ADE}$ & -  & 0.028 & 0.026 & 0.021 & 0.166 & $\bf{0.008}$\\
\hline
\hline
\multirow{5}{*}{$\bf{FDE}$} & $\bf{ZARA}$-$\bf{1}$ & 0.53 & 0.56 & $\bf{0.42}$ & 2.64  & 0.47 \\
\cline{2-7}
~ & $\bf{ZARA}$-$\bf{2}$ & 0.65 & 0.70 & 0.54 & 4.75 & $\bf{0.53}$\\
\cline{2-7}
~ & $\bf{UNIV}$ & 0.77 & 0.84 & 0.75 & 5.95 & $\bf{0.40}$\\
\cline{2-7}
~ & $\bf{ETH}$ & 1.45 & 1.48 & 1.22 & 5.78 & $\bf{0.54}$ \\
\cline{2-7}
~ & $\bf{HOTEL}$ & 1.17 & 1.01 & 0.95 & 4.94 &  $\bf{0.91}$\\
\hline
$\bf{Average\ FDE}$ & -  & 0.91 & 0.91 & 0.78 & 4.81 & $\bf{0.57}$ \\
\hline
$\bf{Variance\ of\ FDE}$ & -  & 0.118 & 0.101 & 0.802 & 1.394 & $\bf{0.031}$\\
\hline
\end{tabular}
\end{table*}

\begin{table*}[h]
\caption{Comparison of Computational Time}
\label{table:3}
\begin{center}
\renewcommand\arraystretch{1.3}
\begin{tabular}{p{4.7cm}<{\centering}|p{1.5cm}<{\centering}|p{2.0cm}<{\centering}|p{2.0cm}<{\centering}|p{2.0cm}<{\centering}|p{2.5cm}<{\centering}}
\hline
$\bf{Metric}$ & $\bf{LSTM}$ & $\bf{Social\ LSTM}$ & $\bf{Social\ GAN}$ & $\bf{Social\ Attention}$ & $\bf{StarNet\ (Ours)}$ \\
\hline
$\bf{Inference\ Time\ (Seconds)}$ & $\bf{0.029}$ & 0.504 & 0.202 & 3.714 & $\bf{0.073}$ \\
\hline
$\bf{Number\ of\ Paramters\ (Kilo)}$ & $\bf{22.87}$ & 156.06 & 108.03 & 874.95 & $\bf{31.90}$ \\
\hline
\end{tabular}
\end{center}
\end{table*}

\section{EXPERIMENTS}
\label{section:4}
We evaluate our model on two human crowded trajectory datasets: ETH \cite{c24} and UCY \cite{c25}. These datasets have 5 sets with 4 different scenes. In these scenes, there exist challenging interactions, such as walking side by side, collision avoidance and changing directions. Following the settings in \cite{c11,c13,c14}, we train our model on 4 sets and test on the remaining one. 

We compare our StarNet with three state-of-the-arts including Social LSTM, Social GAN and Social Attention. Besides, we test the basic LSTM-based encoder-decoder model, which does not consider the interactions among pedestrians, as a baseline. 

Following \cite{c11,c13,c14}, we compare these methods in terms of the Average Displacement Error (ADE) and Final Displacement Error (FDE). The ADE is defined as the mean Euclidean distance between predicted coordinates and the ground truth. Specifically, all methods output 8 coordinates uniformly sampled from the predicted trajectory. Then the distance between such 8 points with the ground truth is accumulated as the ADE. The FDE is the distance between the final point of the predicted trajectory and the final point of the ground truth. All these methods are trained with the loss Eq. (10) to deal with multimodal distribution during evaluation. Besides, we compare the computational time of all these methods. All experiments are conducted on the same computational platform with an NVIDIA Tesla V100 GPU.

\subsection{Experimental Results}

\subsubsection{Accuracy}
As shown in TABLE \ref{table:2}, StarNet outperforms the others in most cases. A possible explanation is that StarNet considers the collective influence among pedestrians all together to make more accurate predictions. In comparison, other state-of-the-arts only model the pairwise interactions between pedestrians.

Interestingly, we notice that the test datasets include multiple senses. In these scenes, StarNet has the smallest variances of ADE and FDE, which means that StarNet is robust against the changes of scenes.

\begin{figure*}[thpb]
\centering
\subfigure[Scene 1] {
\label{Fig.sub.1}
\includegraphics[width=5.5cm]{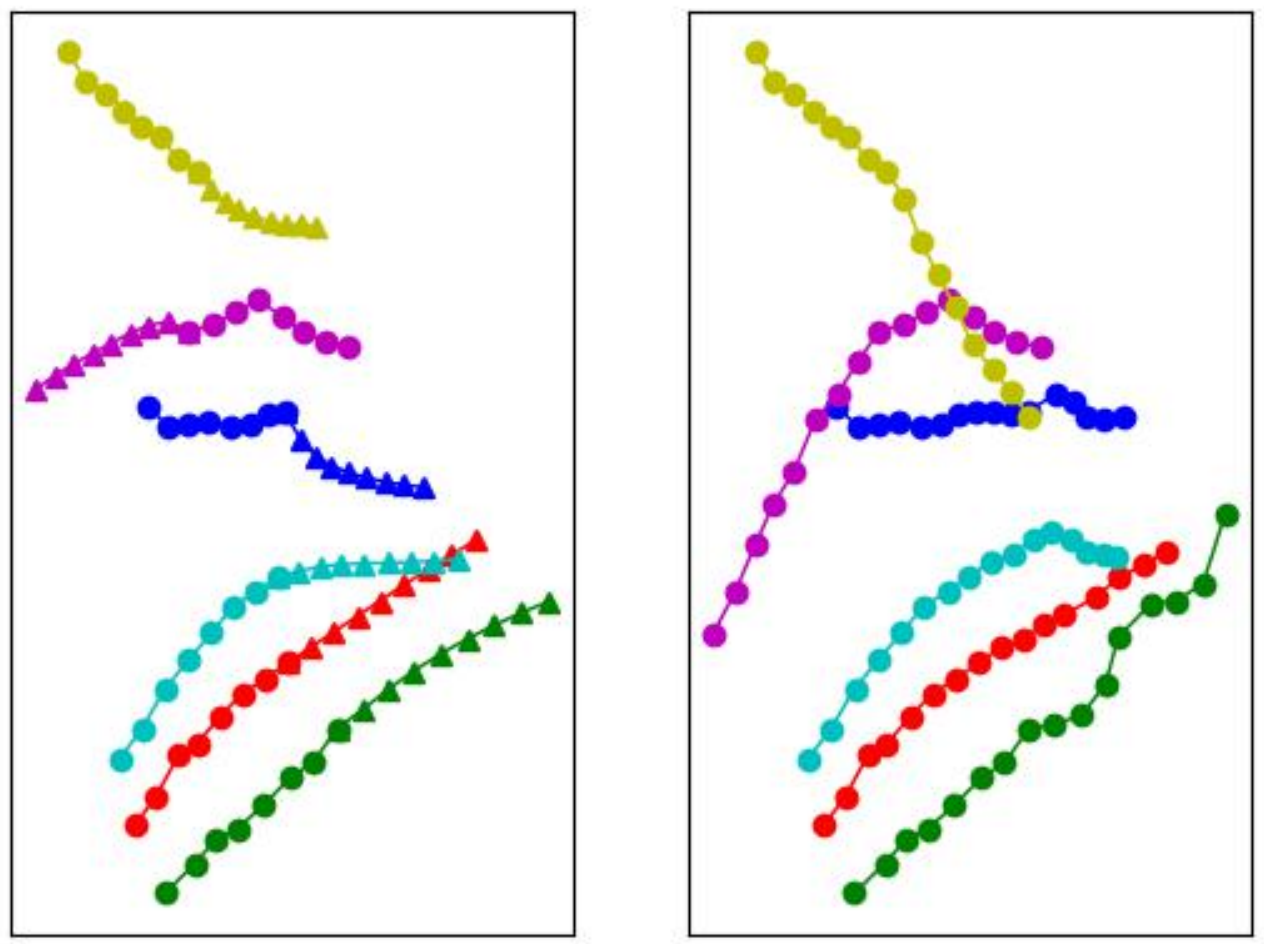}
}
\subfigure[Scene 2] {
\label{Fig.sub.2}
\includegraphics[width=5.5cm]{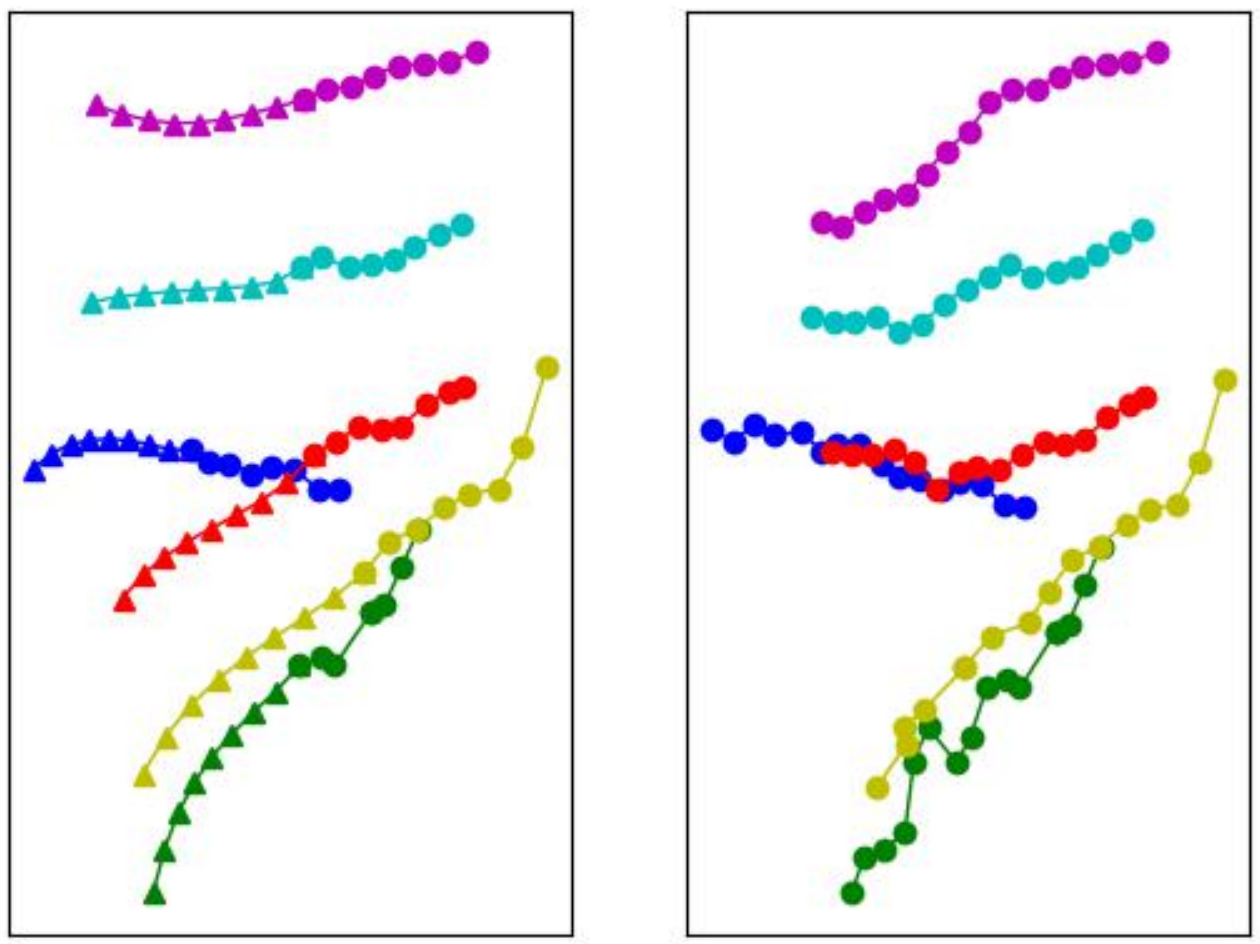}
}

\subfigure[Scene 3] {
\label{Fig.sub.3}
\includegraphics[width=5.5cm]{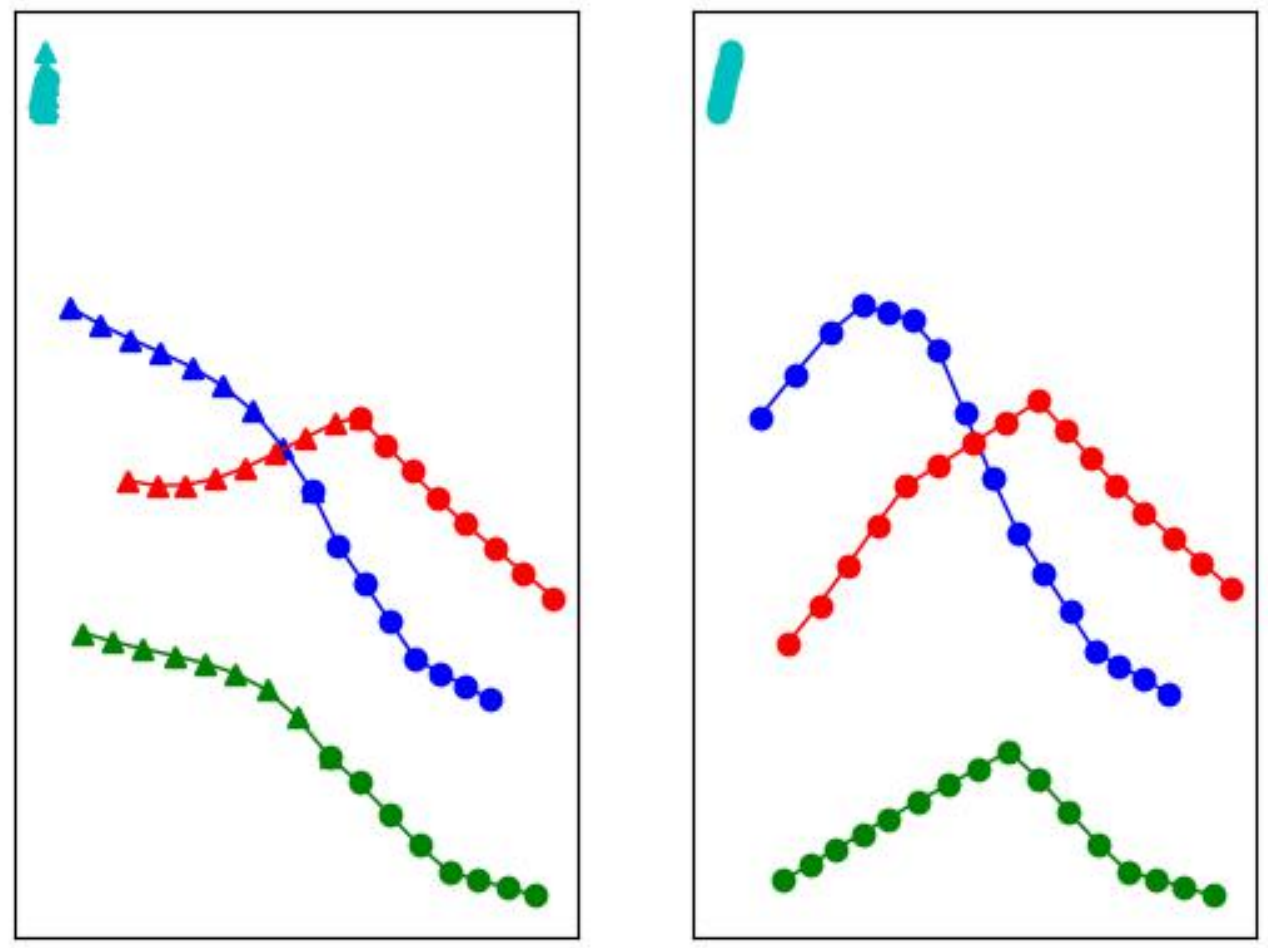}
}
\subfigure[Scene 4] {
\label{Fig.sub.4}
\includegraphics[width=5.5cm]{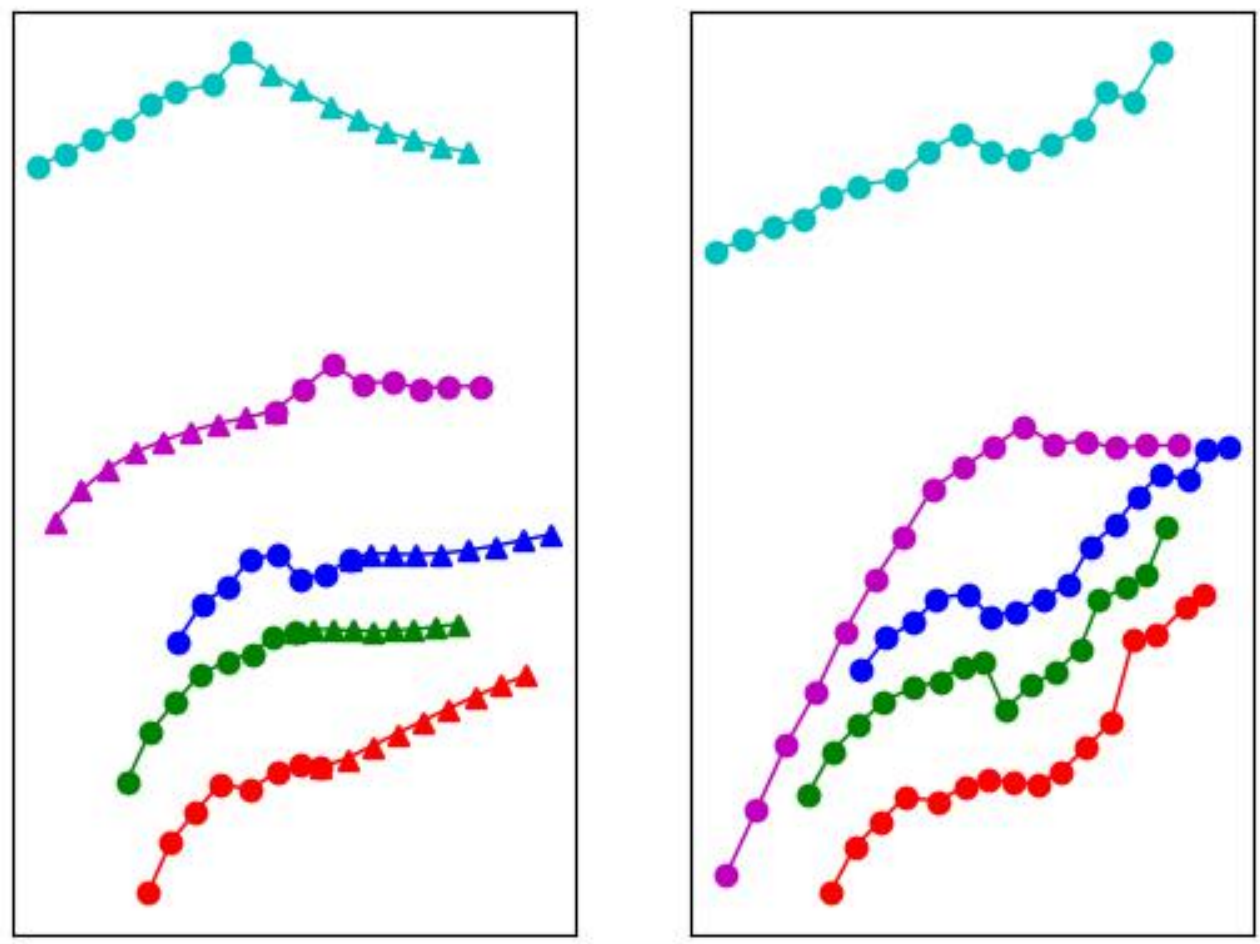}
}

\caption{Predicted trajectories and the corresponding ground truths. Different colors indicate different trajectories. The trajectories of ground truth are labeled with dots. The predicted trajectories are labeled with triangles.}
\label{fig:3}
\end{figure*}

To assess StarNet qualitatively, we illustrate the prediction results in 4 scenes, as shown in Figure \ref{fig:3}. In each scene, the left sub-figure presents the observed trajectories and the predicted trajectories of all pedestrians. The right sub-figure shows the trajectories of ground truth.

We can observe that StarNet could handle complicated interactions among pedestrians. Most predicted trajectories accurately reflect the pedestrians' movements and have no collisions with other trajectories. However, there are some failure cases due to the multimodal distribution of future trajectories. For example, in \ref{Fig.sub.3}, the predictions for the blue and green trajectories fail to match the ground truth. We argue that although these predicted trajectories do not match the ground truth, these trajectories are still plausible in crowded scenes. 


\subsubsection{Computational time cost}
When deployed in mobile robots and autonomous vehicles, the prediction algorithm needs to be invoked with a high frequency. Hence the computational time of the prediction algorithm is a crucial property.

As shown in TABLE \ref{table:3}, the basic LSTM model is the fastest model since the model takes no interactions among pedestrians into consideration. StarNet is the second fastest model. Specifically, StarNet is 51 times faster than Social Attention, 7 times faster than Social LSTM, and 3 times faster than Social GAN. Meanwhile, the number of parameters employed by StarNet is less than state-of-the-arts by a large margin. StarNet is computationally efficient since the interpersonal interactions among pedestrians are computed in a single forward propagation, as discussed in Section \ref{section:2}.

\section{CONCLUSION}
\label{section:5}
In this paper, we propose StarNet, which has a star topology, for pedestrian trajectory prediction. StarNet learns complicated interpersonal interactions and predicts future trajectories with low time complexity. We apply a centralized hub network to model the spatio-temporal interactions among pedestrians. Then the host network takes full advantage of the spatio-temporal representation and predicts pedestrians' trajectories. We demonstrate that StarNet outperforms state-of-the-arts in multiple experiments.


\begin{thebibliography}{99}

\bibitem{c1} D. Ferguson, D. Michael, U. Chris and K. Sascha, ``Detection, prediction, and avoidance of dynamic obstacles in urban environments," \textit{in 2008 IEEE International Conference on Intelligent Vehicles Symposium (IVS)}. IEEE, 2008, pp. 1149-1154.
\bibitem{c2} Y. Luo, P. Cai, A. Bera, D. Hsu, W. S. Lee and D. Manocha, ``Porca: Modeling and planning for autonomous driving among many pedestrians," \textit{IEEE Robotics and Automation Letters}, vol. 3, no. 4, pp. 3418-3425, 2018.
\bibitem{c3} F. Large, D. Vasquez, T. Fraichard and C. Laugier, ``Avoiding cars and pedestrians using velocity obstacles and motion prediction," \textit{in 2004 IEEE International Conference on Intelligent Vehicles Symposium (IVS)}. IEEE, 2004, pp. 375-379.
\bibitem{c4} B. D. Ziebart, N. Ratliff, G. Gallagher, C. Mertz, K. Peterson, J. A. Bagnell, M. Hebert, A. K. Dey and S. Srinivasa, ``Planning-based prediction for pedestrians," \textit{in 2009 IEEE/RSJ International Conference on Intelligent Robots and Systems (IROS)}. IEEE, 2009, pp. 3931-3936.
\bibitem{c5} P. Trautman and A. Krause, ``Unfreezing the robot: Navigation in dense, interacting crowds," \textit{in 2010 IEEE/RSJ International Conference on Intelligent Robots and Systems (IROS)}. IEEE, 2010, pp. 797-803.
\bibitem{c6} N. E. D. Toit and J. W. Burdick, ``Robot motion planning in dynamic, uncertain environments," \textit{IEEE Transactions on Robotics}, vol. 28, no. 1, pp. 101-115, 2012.
\bibitem{c7} D. Helbing and P. Molnar, ``Social force model for pedestrian dynamics," \textit{Physical review E}, vol. 51, no. 5, pp. 4282, 1995.
\bibitem{c8} S. Yi, H. Li and X. Wang, ``Understanding pedestrian behaviors from stationary crowd groups," \textit{in 2015 IEEE Conference on Computer Vision and Pattern Recognition (CVPR)}. IEEE, 2015, pp. 3488-3496.
\bibitem{c9} S. Yi, H. Li and X. Wang, ``Pedestrian behavior modeling from stationary crowds with applications to intelligent surveillance," \textit{IEEE transactions on image processing}, vol. 25, no. 9, pp. 4354-4368, 2016.
\bibitem{c10} B. Zhou, X. Wang and X. Tang, ``Understanding collective crowd behaviors: Learning a mixture model of dynamic pedestrian-agents," \textit{in 2012 IEEE Conference on Computer Vision and Pattern Recognition (CVPR)}. IEEE, 2012, pp. 2871-2878.
\bibitem{c11} A. Alahi, K. Goel, V. Ramanathan, A. Robicquet, F. Li and S. Savarese, ``Social lstm: Human trajectory prediction in crowded spaces," \textit{in 2016 IEEE Conference on Computer Vision and Pattern Recognition (CVPR)}. IEEE 2016, pp. 961-971.
\bibitem{c12} H. Wu, Z. Chen, W. Sun, B. Zheng and W. Wang, ``Modeling trajectories with recurrent neural networks," \textit{in 28th International Joint Conference on Artificial Intelligence (IJCAI)}. 2017, pp. 3083-3090.
\bibitem{c13} A. Gupta, J. Johnson, F. Li, S. Savarese and A. Alahi, ``Social GAN: Socially acceptable trajectories with generative adversarial networks," \textit{in 2018 IEEE Conference on Computer Vision and Pattern Recognition (CVPR)}. IEEE, 2018, pp. 2255-2264.
\bibitem{c14} A. Vemula, K. Muelling and J. Oh, ``Social attention: Modeling attention in human crowds," \textit{in 2018 IEEE International Conference on Robotics and Automation (ICRA)}. IEEE, 2018, pp. 1-7.
\bibitem{c15} Y. Xu, Z. Piao and S. Gao S, ``Encoding crowd interaction with deep neural network for pPedestrian trajectory prediction," \textit{in 2018 IEEE Conference on Computer Vision and Pattern Recognition (CVPR)}. IEEE, 2018, pp. 5275-5284.
\bibitem{c16} I. Goodfellow, J. Pouget-Abadie, M. Mirza, B. Xu, D. Warde-Farley, S. Ozair, A. Courville and Y. Bengio, ``Generative adversarial nets," \textit{in 28th Conference on Neural Information Processing Systems (NIPS)}. 2014, pp. 2672-2680.
\bibitem{c17} C. M. Bishop, ``Mixture density networks," \textit{Technical Report NCRG/4288}, Aston University, Birmingham, UK, 1994.
\bibitem{c18} D. Ha and D. Eck, ``A neural representation of sketch drawings," \textit{arXiv preprint} arXiv:1704.03477, 2017.
\bibitem{c19} E. Schmerling, K. Leung, W. Vollprecht and M. Pavone, ``Multimodal probabilistic model-based planning for human-robot interaction," \textit{in 2018 IEEE International Conference on Robotics and Automation (ICRA)}. IEEE, 2018, pp. 1-9.
\bibitem{c20} K. Cho, B. V. Merrienboer, C. Gulcehre, D. Bahdanau, F. Bougares, H. Schwenk and Y. Bengio, ``Learning phrase representations using RNN encoder-decoder for statistical machine translation," \textit{arXiv preprint} arXiv:1406.1078, 2014.
\bibitem{c21} K. Cho, B. V. Merrienboer, D. Bahdanau and Y. Bengoi, ``On the properties of neural machine translation: Encoder-decoder approaches," \textit{arXiv preprint} arXiv:1409.1259, 2014.
\bibitem{c22} D. Bahdanau, J. Chorowski, D. Serdyuk, P. Brakel and Y. Bengio, ``End-to-end attention-based large vocabulary speech recognition," \textit{in 2015 International Conference on Acoustics, Speech and Signal Processing (ICASSP)}. IEEE, 2016, pp. 4945-4949.
\bibitem{c23} C. R. Qi, H. Su, K and J. G. Leonidas, ``Pointnet: Deep learning on point sets for 3d classification and segmentation," \textit{in 2017 Conference on Computer Vision and Pattern Recognition (CVPR)}, IEEE, 2017, pp. 652-660.
\bibitem{c24} S. Pellegrini, A. Ess, K. Schindler and L. V. Gool, ``You'll never walk alone: Modeling social behavior for multi-target tracking," \textit{in 2009 IEEE International Conference on Computer Vision (ICCV)}. IEEE, 2009, pp. 261-268.
\bibitem{c25} A. Lerner, Y. Chrysanthou and D. Lischinski, ``Crowds by example," \textit{Computer Graphics Forum}, vol. 26, no. 3, pp. 655-664, 2007.
\bibitem{c26} D. Varshneya, G. Srinivasaraghavan, ``Human trajectory prediction using spatially aware deep attention models," \textit{arXiv preprint} arXiv:1705.09436, 2017.
\bibitem{c27} T. Fernando, S. Denma, S. Sridharan and C. Fookes, ``Soft+hardwired attention: An lstm framework for human trajectory prediction and abnormal event detection," \textit{arXiv preprint} arXiv:1702.05552, 2017.

\end{thebibliography}
\end{document}